\documentclass[letterpaper, 10 pt, conference]{ieeeconf}  % Comment this line out if you need a4paper

\IEEEoverridecommandlockouts                              % This command is only needed if 
\usepackage{amsmath}
\usepackage{graphicx}
\usepackage{algorithm}
\usepackage{algpseudocode}
\usepackage{textcomp}
\usepackage{xcolor}
\usepackage{multicol}
\usepackage{tabularx}
\usepackage{rotating}
\usepackage{array}
\usepackage{cite}
\usepackage{booktabs}   % 美观表格横线
\usepackage{makecell} 
\usepackage{float}
\usepackage{stfloats}
\usepackage{placeins}
\usepackage{mathrsfs}
\usepackage{amsfonts}
\usepackage{adjustbox}
\let\labelindent\relax
\usepackage{enumitem}
\usepackage{multirow}
\usepackage[hyphens]{url}
\usepackage{hyperref}

\def\BibTeX{{\rm B\kern-.05em{\sc i\kern-.025em b}\kern-.08em
    T\kern-.1667em\lower.7ex\hbox{E}\kern-.125emX}}
\title{\LARGE \bf
Vision-Based Obstacle Separation for Strawberry Harvesting in Clusters Using Hierarchical Reinforcement Learning
}

\author{Teng Li$^1{^{,2}}$, Hanfei Shi$^{3}$, Chunjiang Zhao$^{1}$* and Ya Xiong$^{1}$*% <-this % stops a space
\thanks{This work was supported by the Innovation Ability Project of Beijing Academy of Agricultural and Forestry Sciences (BAAFS) (KJCX20240321), the Outstanding Youth Foundation of BAAFS (YKPY2025007), the Haidian District Bureau of Agriculture and Rural Affairs, and the NSFC Excellent Young Scientists Fund (overseas). (*\textit{Corresponding Author: Chunjiang Zhao \tt\small zhaocj@nercita.org.cn; Ya Xiong, \tt\small yaxiong@nercita.org.cn})}% <-this % stops a space
\thanks{$^1$The Intelligent Equipment Research Center, Beijing Academy of Agriculture and Forestry Sciences, Beijing 100097, China.}%
\thanks{$^{2}$School of Intelligence Science and Technology, University of Science and Technology Beijing, Beijing 100083, China}%
\thanks{$^3$Happy Elements Ltd., Beijing 100094, China.}%
}

\begin{document}
\raggedbottom

\maketitle
\thispagestyle{empty}
\pagestyle{empty}

%%%%%%%%%%%%%%%%%%%%%%%%%%%%%%%%%%%%%%%%%%%%%%%%%%%%%%%%%%%%%%%%%%%%%%%%%%%%%%%%
\begin{abstract}

Selective harvesting in clustered strawberry environments is challenging because ripe fruits are often occluded by surrounding unripe fruits, making direct grasping unreliable. To address this problem, this paper proposes a hierarchical reinforcement learning framework, termed VGPA, which integrates a vision-guided decision mechanism and a Progressive Adaptive Exploration Strategy (PAES) for vision-based obstacle separation and harvesting. The task was decomposed into two sequential stages: obstacle separation and target grasping. At the high level, the vision-guided mechanism improved option selection and accelerated policy convergence. At the low level, PAES improved exploration efficiency and training stability during continuous control learning. In simulation experiments, the learned policy achieved a success rate of 96.7\%. In addition, sim-to-real transfer experiments on a self-developed parallel robot showed that the proposed method achieved success rates ranging from 71.7\% to 88.3\%, outperforming direct picking while requiring only 1.22~s more average harvesting time. These results verified the effectiveness, generalization ability, and practical potential of the proposed method for robotic harvesting in complex clustered environments.

\end{abstract}

%%%%%%%%%%%%%%%%%%%%%%%%%%%%%%%%%%%%%%%%%%%%%%%%%%%%%%%%%%%%%%%%%%%%%%%%%%%%%%%%

\section{INTRODUCTION}

In recent years, the demand for automated fruit and vegetable harvesting has grown rapidly with the increasing demand for labor-saving agricultural production \cite{meshram2021machine}. For cluster-growing crops such as strawberries, tomatoes, grapes, and cherries, harvesting requires selectively picking ripe fruits without damaging surrounding unripe ones \cite{ge2019fruit}. Achieving reliable selective harvesting has therefore become a key challenge in agricultural robotics.

\begin{figure}[htbp!]
    \centering
    \includegraphics[width=0.4\textwidth]{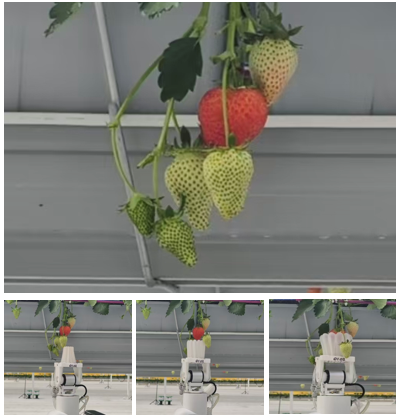}
    \caption{Top: A target strawberry cluster in the initial observation, where the target strawberry is surrounded by obstacles. Bottom (left to right): The robot learns to push away the obstacles and finally grasp the target.}
    \label{fig:overall}
\end{figure}
Most existing studies addressed this problem through obstacle avoidance. For example, in tomato harvesting, deep reinforcement learning combined with spatial pose modeling has been used to determine an appropriate grasp posture for the end-effector \cite{li2024peduncle}. In guava harvesting, obstacle detection and shape approximation were combined with an LSTM-based DDPG method to generate collision-free paths in configuration space \cite{yandun2021reaching}. These approaches were effective when obstacles were sparse and sufficient free space is available. However, in real clustered harvesting environments, ripe fruits were often tightly surrounded by unripe fruits or foliage, leaving little room for collision-free access. As a result, passive obstacle avoidance alone is often insufficient. As illustrated in Fig.~\ref{fig:overall}, the target strawberry is frequently enclosed by nearby obstacles, making direct grasping difficult and motivating the need for obstacle separation before harvesting.

To overcome this limitation, active obstacle separation can be used to push or separate obstacles before grasping. Compared with simple obstacle avoidance, this strategy allowed the robot to access targets that would otherwise be unreachable. Xiong et al. \cite{xiong2021improved} proposed a 3D vision-based obstacle separation strategy that combined horizontal pushing, vertical pushing, and dragging to improve harvesting success. Williams et al. \cite{williams2025zero} introduced a DRM-based deep reinforcement learning framework integrating Cartesian impedance control and a custom Mujoco simulation environment to reduce the sim-to-real gap.

Although active obstacle separation has shown promise, clearance and grasping form an interdependent multi-stage process, and coordinating them efficiently remains challenging. Hierarchical reinforcement learning (HRL) is well suited to such long-horizon sequential tasks. Xu \cite{xu2021efficient} proposed a goal-conditioned HRL framework for push-and-grasp coordination in cluttered environments. Ren \cite{ren2023learning} introduced a dual-function push-and-grasp strategy within a hierarchical framework. Yang et al. \cite{yang2024learning} employed HRL to combine learned parameterized manipulation primitives and adopted a two-stage training process to improve both precise grasping and push-and-grasp coordination.

To address these challenges, this paper proposes an obstacle-separation harvesting method based on hierarchical reinforcement learning. When the target strawberry was occluded, the robotic arm first cleared the obstructing fruits and then performed stable grasping of the target. Based on this idea, we developed the VGPA framework to improve the efficiency and robustness of clustered strawberry harvesting. The main contributions are as follows:

\begin{itemize}
    \item
% An active obstacle separation strategy based on deep reinforcement learning was proposed, enabling robots to proactively remove obstructions in clustered fruit environments.
A vision-guided hierarchical reinforcement learning framework was developed to coordinate obstacle separation and target grasping for long-horizon strawberry harvesting tasks under severe occlusion.

    \item
% A hierarchical reinforcement learning framework, VGPA, was constructed to coordinate obstacle separation and harvesting, thereby improving robustness and stability in complex tasks.
A progressive adaptive exploration strategy (PAES) was introduced to dynamically regulate low-level exploration according to the learning progress, improving the stability and efficiency of hierarchical policy learning.

    \item
% A simulated environment covering diverse real-world occlusion scenarios was established, in which deep reinforcement learning agents were trained to adapt to varying occlusion conditions and achieve improved generalization capability.
The proposed framework achieved robust sim-to-real transfer and consistently outperformed direct picking under different occlusion conditions.
\end{itemize}
\section{RELATED WORK}

\subsection{Option learning}

Option learning represents a classical paradigm in reinforcement learning, which enhances exploration and exploitation by training sub-policies under sparse reward settings. Directly addressing complex and long-horizon tasks remains highly challenging, thus making option-based approaches particularly appealing in robotic learning. Sutton et al. \cite{sutton1999between} introduced the option framework, which consisted of an initial set, intra-option policies, and termination conditions. This framework abstracted policy learning into a hierarchical structure, where high-level policies were responsible for selecting options while low-level policies executed primitive actions. Building on the actor–critic paradigm, Bacon et al. \cite{bacon2017option} proposed the option-critic architecture, which employed gradient-based optimization to jointly learn option policies, termination functions, and value functions across two temporal scales. This automated option discovery enabled efficient handling of complex problems. Following this, a variety of option-based reinforcement learning algorithms have been developed, including option-driven methods \cite{zhang2021hierarchical} and the universal option framework  \cite{yang2021hierarchical}. Nevertheless, despite these advances, option learning still suffers from challenges such as non-stationarity and limited efficiency.
The primary problem addressed in this study was related to HRL. For the long-term sequence problem of strawberry picking, where occlusions complicated grasping, HRL demonstrated considerable potential. However, HRL itself suffers from two inherent issues: difficulty in converging high-level policies during parallel training, and challenges in learning low-level tasks and achieving generalization.

\subsection{Push and Grasp}
 Certain fruits and vegetables (e.g., strawberries, tomatoes) often grow in clusters, where ripe fruits may be obscured by unripe fruits or foliage. Single-grasp operations struggled to harvest without damaging unripe fruits; thus, pre-grasp clearance actions like pushing were required to expose and make targets accessible. Push-and-grasp constituted a long-sequence decision and control problem: Xu et al. \cite{xu2021efficient} proposed a goal-conditioned hierarchical reinforcement learning framework for learning push-and-grasp cooperative policies in cluttered environments. Ren et al. \cite{ren2023learning} designed a dual-function strategy that efficiently coordinated pushing and grasping actions to address both goal-independent and goal-directed tasks.
 Wang et al. \cite{wang2025hierarchical} employed the vision-based hierarchical coordination strategy
 HCLM for coordinating specific action primitives. However, these methods\cite{tahmaz2025impedance}, \cite{hoang2022context}, \cite{zhou2023learning}  were predominantly validated in constrained tabletop environments, posing challenges for direct transfer to agricultural settings. 

Fruit distribution is three-dimensional, supports and constraints are weak, plants are deformable, and safety constraints are stringent. Selective harvesting requires tight coupling with subsequent three-dimensional path planning. HRL provides a natural paradigm for addressing agricultural tasks that require both obstacle separation and grasping. Within this framework, the high-level policy was responsible for temporal decision-making and sub-goal selection (i.e., deciding when to clear and when to grasp), while the low-level controllers specialized in executing safe pushing actions for obstacle separation and robust grasping strategies for fruit harvesting. This hierarchical decomposition not only reduced the complexity of learning in highly cluttered environments but also better accommodated the intrinsic requirement of selective harvesting—performing proactive obstacle separation before executing a stable grasp.

\begin{figure*} 
\centering 
\includegraphics[width=1.0\textwidth]{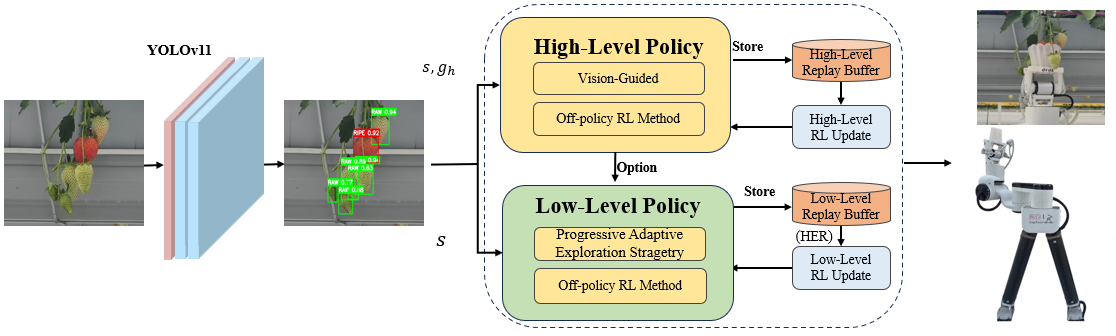} 
\caption{Overall architecture of the proposed VGPA framework. The RGB-D observation was first processed by YOLOv11 to detect and localize the target strawberry and surrounding obstacles. Based on the perceived occlusion state, the high-level policy selected an option (obstacle separation or grasping) with the assistance of the vision-guided module. The low-level policy then generated continuous control actions using PAES and DDPG, while HER improved sample efficiency through goal relabeling. The resulting actions were executed by the robot to perform obstacle separation and target grasping, and the interaction data were stored for hierarchical policy training.} 
\label{fig:myfigure} 
\end{figure*}

\section{Method}
To address the clustered strawberry harvesting task, we proposed VGPA, a hierarchical reinforcement learning framework that combines visual guidance with progressive exploration, as illustrated in Fig.~\ref{fig:myfigure}. In the proposed framework, the robot interacted with the environment through a hierarchical goal-conditioned policy. The high-level policy planned the task by selecting options according to the current environmental state and the high-level goal, where each option corresponded to a low-level sub-objective. The low-level policy then executed continuous robotic actions based on environmental observations and the assigned sub-objective. As the low-level policy interacted with the environment, the state was updated step by step, and once the current sub-objective was achieved, the high-level policy issued a new option. The resulting high-level and low-level transitions were stored in their corresponding replay buffers for training.

During training, mini-batches of transitions were sampled from the replay buffers to update the model parameters. The high-level policy was optimized using DIOL, while the low-level policy was trained using DDPG \cite{lillicrap2015continuous}. In addition, HER was employed at the low level to improve sample efficiency in goal-conditioned continuous control. At the high level, visual guidance accelerated option learning and improved decision efficiency. At the low level, PAES was introduced to improve exploration efficiency and stabilize hierarchical training by adaptively adjusting the exploration behavior according to the training progress. The overall training procedure of VGPA was summarized in Algorithm~\ref{alg:VGPA}.

\begin{algorithm}[htbp!]
\caption{VGPA: Vision-Guided Progressive Adaptive Exploration Strategy}
\label{alg:VGPA}
\begin{algorithmic}[1] 
\For{$\text{epoch}=1$ \textbf{to} $M1$}
     \For{$\text{episode}=1$ \textbf{to} $M2$}
         \State Sample a high-level goal $g^H$
         \If{current episode $\in$ VG episodes}
             \State Obtain an option $o_t$ as a low-level subgoal
         \Else
             \State Sample an option $o_t$ as a low-level subgoal
         \EndIf
         \For{$\text{step}=1$ \textbf{to} $M3$}
             \State Sample a low-level goal $g^L$ 
             \State from the high-level policy $\pi^H$
             \While{\textbf{not} low-level subgoal achieved}
                 \State Sample $a^L \sim \pi^L(\cdot\mid s,g^L)$ with PAES
                 \State Execute $a^L$ and observe $s'$
                 \State Store transition $(s, a^L, r, s', g^L)$
                 \State $s \gets s'$
             \EndWhile
             \ForAll{transitions $(\tilde s,\tilde a^L,\tilde r,\tilde s', g^L)$ 
                 \State  in the episode}
                 \State Sample hindsight goal $g^L_f$ from achieved 
                 \State states
                 \State Store $(\tilde s,\tilde a^L,\tilde r',\tilde s', g^L_f)$ with HER
             \EndFor
         \EndFor
         \State Perform optimization steps with DDPG
         \State Perform optimization steps with DIOL
     \EndFor
     \State Update the parameters of VG and PAES according to Eqs. (3) and (4).
\EndFor
\end{algorithmic}
\end{algorithm}

\subsection{Image processing}

The image processing included the detection and localization of ripe strawberries and the determination of obstacles near each target. The image processing pipeline consisted of two steps: (1) object detection and localization based on pre-trained deep learning models, and (2) obstacle calculation.
We employed a YOLOv11\cite{sun2026light} object detection network to output rectangular bounding boxes for strawberries along with their classification (ripe/unripe). For each detected bounding box, we calculated the center pixel coordinates and obtained the corresponding depth value from the PyBullet~\cite{coumans2021pybullet} camera's depth map. Subsequently, using the inverse transformation of the camera's projection matrix and view matrix, we back-projected the pixel coordinates into three-dimensional space. This yielded the strawberries' 3D positions within the simulated world coordinate system, which we then used for subsequent robotic arm motion planning.
In obstacle calculation for elevated strawberry harvesting, we defined a region of interest (ROI) around the target strawberry, primarily located below and in its surrounding area—corresponding to the position where the robotic arm performed the swallowing motion. If other strawberries were detected within this ROI, the positional information was incorporated into the state representation.

\subsection{Vision-Guided Module}
To improve the learning efficiency of the high-level policy and alleviate non-stationarity during hierarchical training, we introduced a vision-guided module to assist option selection. This module provided high-level decision guidance by estimating the occlusion status of the target strawberry from visual observations. Specifically, it adjusted the proportion of vision-guided options derived from image processing and options generated by the learned high-level policy.

\begin{figure}[htbp!] 
\centering 
\includegraphics[width=0.4\textwidth]{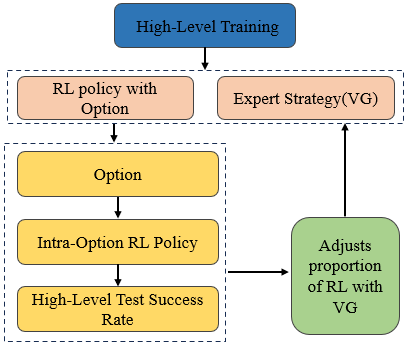} 
\caption{The structure of VG at the high level.} 
\label{fig:VG} 
\end{figure}

During the early stage of training, a larger proportion of vision-guided episodes was used to provide reliable option supervision, thereby reducing non-stationarity and accelerating the convergence of the high-level policy. As the success rate of the high-level policy increases, the proportion of vision-guided episodes was gradually reduced, allowing the policy to rely more on its own learned decisions. The training schedule of the high-level policy is illustrated in Fig.~\ref{fig:VG}. Training was divided into epochs, and each epoch contained $M_{2}$ interaction episodes. The vision-guided module adaptively determined the number of guided episodes according to the learning progress of the high-level policy, thereby balancing vision-guided supervision and RL-based option selection.

\begin{equation}
\resizebox{0.9\linewidth}{!}{$
D=\left\lfloor\left(\alpha_{\max}-(\alpha_{\max}-\alpha_{\min})\cdot\frac{1}{1+\exp(-k\cdot(\tau-\tau_0))}\right)\cdot e^{-\lambda\cdot\mathrm{EMA}(sr_h)}\cdot M_{2}\right\rfloor
$}
\tag{3}
\end{equation}
where, $M_{2}$ denoted the total number of episodes in each epoch. $\alpha_{\max}$ and $\alpha_{\min}$ represented the upper and lower bounds of the demonstration ratio, ensuring that the number of guided episodes remains within a reasonable range. $\tau$ denotes the current epoch index, $\tau_{0}$ is the inflection point of the curriculum schedule, and $k$ controls the steepness of the logistic decay curve. Together, these parameters determine how the reliance on visual guidance decreases progressively as training proceeds. In addition, $sr_{h}$ denoted the success rate of the high-level policy, and $\mathrm{EMA}(sr_{h})$ is its exponentially smoothed value used to reduce stochastic fluctuations. The parameter $\lambda$ scaled the influence of high-level performance, such that a higher success rate accelerates the reduction of vision-guided episodes.

\subsection{Progressive Adaptive Exploration Strategy (PAES)}

% A novel exploration strategy, PAES, was proposed.
An adaptive exploration strategy, PAES, was designed for hierarchical reinforcement learning to regulate exploration behaviors according to the learning progress of different manipulation stages.
As discussed above, when training different levels of policies in parallel within a hierarchical reinforcement learning system, the stochastic exploration behavior of lower-level policies could introduce delayed instability, which resulted in non-stationary transitions in the MDPs of higher-level structures. Previous work typically addressed this issue by training the upper- and lower-level networks separately, keeping the lower-level policy fixed during high-level training. Although effective, such separated training was costly, especially when data collection was expensive (e.g., in real-world tasks or complex simulations). Another representative method, HAC \cite{levy2017learning}, modified higher-level transitions by assuming the lower-level policy was already optimal; however, this did not fundamentally eliminate non-stationarity.

% In this work, PAES was designed as an adaptive exploration strategy for low-level continuous control.
In this work, PAES was designed as a stage-aware adaptive exploration strategy for low-level continuous control, aiming to improve the stability of hierarchical policy learning rather than simply increasing exploration.
Instead of relying on a fixed exploration schedule, it regulated the balance among random exploration, guided action sampling, and policy-driven execution according to the learning progress of each stage.

Specifically, the agent's action selection in PAES consisted of three mechanisms. First, with probability $\alpha_j$, the agent sampled an action uniformly from the action space $\mathcal{A}$, enabling global and aggressive exploration. Second, with probability $\gamma_j$, the agent executed a guiding policy $\pi_g(s)$ learned from a previously mastered stage and perturbed the action with Gaussian noise $\mathcal{N}(0,(\kappa\sigma_j)^2)$, allowing the agent to enter effective state regions more efficiently. Third, with probability $1-\alpha_j-\gamma_j$, the agent executed the current-stage policy $\pi_j(s)$ with Gaussian noise $\mathcal{N}(0,\sigma_j^2)$ added to the action, which supported fine-grained local exploration. The overall behavior policy is defined as follows:
\begin{equation}
\resizebox{0.9\linewidth}{!}{$
 \pi^{PAES}(a|s)=
\begin{cases}
a\sim\mathrm{Uniform}(\mathcal{A}), & \text{ }\alpha_j, \\
 \\
a\sim\pi_g(s)+\mathcal{N}(0,(\kappa\sigma_j)^2), & \text{ }\gamma_j, \\
 \\
a\sim\pi_j(s)+\mathcal{N}(0,\sigma_j^2), & \text{ }1-\alpha_j-\gamma_j. 
\end{cases}
$}
\tag{4}
\end{equation}

During training, PAES adaptively updated its exploration parameters according to the learning status of each stage. The random action probability and noise scale are updated as follows:
\begin{equation}
\resizebox{0.9\linewidth}{!}{$
\alpha_{j,e+1} = c_\alpha \big(1 - S_e[j]\big); \quad
\sigma_{j,e+1} = c_\sigma \big(1 - S_e[j]\big),
$}
\tag{5}
\end{equation}
where, $S_e[j]$ denotes the smoothed success rate of stage $j$ after the $e$-th training epoch, and $c_\alpha, c_\sigma \in (0,1]$ are upper-bound constants. Meanwhile, the guiding probability was determined by the performance gap between the previously mastered stage and the current stage:
\begin{equation}
\resizebox{0.9\linewidth}{!}{$
\gamma_j = \text{clip}\Big(c_g \cdot \big(S_e[g]-S_e[j]\big)_+, \, \gamma_{\min}, \, \gamma_{\max}\Big),
$}
\tag{6}
\end{equation}
where, $g$ denotes the highest mastered stage index, and $\gamma_{\min}, \gamma_{\max}$ are the lower and upper bounds of the guiding probability.

Through this adaptive mechanism, PAES gradually reshaped the exploration behavior throughout training. At the beginning of training, when the success rates were close to zero, large random action probability and noise variance were assigned to ensure sufficient global exploration. During intermediate stages, when a previous stage has been mastered but the current stage remains unstable, the guiding probability increased, enabling the agent to make greater use of effective prior policies and reach high-value regions more efficiently. As training converges and the success rate approached one, $\alpha_j$, $\sigma_j$, and $\gamma_j$ gradually decreased toward zero, so the agent increasingly relied on stable policy execution. In this way, PAES improved sample efficiency and training stability by adapting the action-sampling behavior during training, while keeping the original reward definition and optimization objective unchanged.

\subsection{Reward Function}
Since our task was a two-stage obstacle separation and grasping task, we represented high-level goals as two-dimensional vectors$\mathbf{g}^{\mathcal{H}}=( \begin{array} {c}{{\mathbf{g}_{1}^{\mathcal{L}}},{\mathbf{g}_{2}^{\mathcal{L}}}} \end{array})$. The vector (1,0) indicated that the low-level goal $\mathbf{g}_1^{\mathcal{L}}$ should be achieved, while (0, 1) indicated that the low-level goal $\mathbf{g}_2^{\mathcal{L}}$ should be achieved. The high-level reward function was defined simply by checking whether the desired high-level goal was achieved, using element-wise equality with the actual achieved goal $\mathbf{g}^{\mathcal{h}}$, as shown in the following equation.

\begin{equation}
\left.r^{\mathcal{H}}=\left\{
\begin{array}
{ll}0, & \mathbf{g}^{\mathcal{H}}=\mathbf{g}^{\mathcal{h}} \\
-1, & \mathbf{g}^{\mathcal{H}}\neq\mathbf{g}^{\mathcal{h}}
\end{array}\right.\right.
\tag{7}
\end{equation}

In low-level tasks, the reward function comprised distance and pose rewards, combined through weighting to simultaneously ensure precision and stability of movements. Both employed sparse rewards: when the distance was below the set threshold, the reward was 0; otherwise, it was -1. A reward of 0 was given when the end-effector's orientation maintained an angle within ±10° relative to the vertical direction, and -1 otherwise, ensuring the gripper maintains an appropriate posture. In summary, the low-level reward was a weighted sum of two components: the distance reward and the posture reward, weighted at 0.7 and 0.3 respectively. This effectively constrained and optimized the low-level pushing and grasping actions.
\begin{equation}
\mathrm{r}_p=
\begin{cases}
0, & 0.04\leq (obs-to-tar)_x\leq 0.045 \\
-1, & \mathrm{otherwise}
\end{cases}
\tag{8}
\end{equation}

\begin{equation}
\mathrm{r}_g=
\begin{cases}
0, & 0.0\leq (ee-to-tar)\leq 0.01 \\
-1, & \mathrm{otherwise}
\end{cases}
\tag{9}
\end{equation}

\begin{equation}
r_{\mathrm{pose}}(\theta)=
\begin{cases}
0, & |\theta|\leq10^\circ \\
-1, & \mathrm{otherwise} 
\end{cases}
\tag{10}
\end{equation}

\section{EXPERIMENTS}

\subsection{Experiment setup and training}
 To evaluate the efficacy of the proposed method, this paper conducted simulation experiments within the PyBullet \cite{coumans2021pybullet} framework based on the Bullet engine. The simulation environment was established by modeling strawberry racks and strawberries in SolidWorks and then exporting them into the simulator. Multiple representative occlusion scenarios were constructed, including cases where the target fruit was occluded by a single unripe fruit and cases where it was simultaneously occluded by multiple fruits, in order to systematically evaluate the performance of the proposed method under different levels of task complexity. The high-level and low-level reinforcement learning policies were implemented using DIOL and DDPG, respectively. Both the actor and critic networks adopted fully connected architectures, with ReLU as the activation function and Adam as the optimizer. The learning rate was set to $1\times10^{-3}$. The batch size for experience replay was 256, and the replay buffer size was $1\times10^{6}$. The model was trained for 300 epochs, with each epoch consisting of 100 episodes and each episode containing 200 steps. Training was conducted over 100 distinct scenarios to improve the robustness and generalization ability of the learned policy.

\begin{figure}[t] 
\centering 
\includegraphics[width=0.48\textwidth]{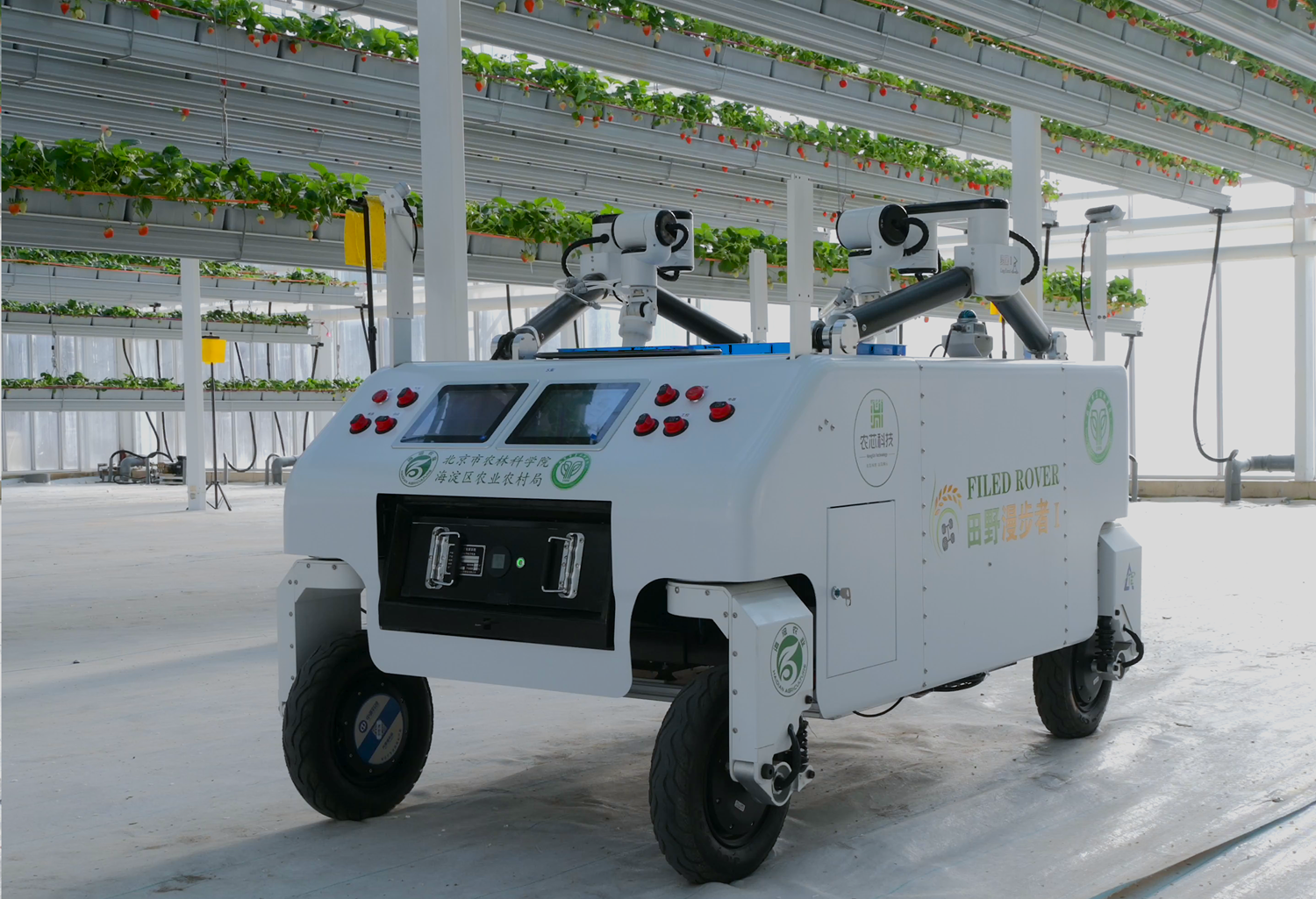} 
\caption{Field test of the strawberry harvesting robot (HarvestFlex) in a greenhouse environment (Beijing Cuihu Agriculture Technology Co., Ltd.).} 
\label{fig:field_test} 
\end{figure}

Based on the simulation validation, real-world experiments were further carried out to evaluate the transferability of the learned policy in practical settings. The real-world platform employed a hybrid robotic arm (LingXtend)\cite{chen2024design} to perform strawberry harvesting under occlusion. As shown in Fig.~\ref{fig:field_test}, the developed strawberry harvesting robot was deployed for on-site validation in the glasshouse of Cuihu Agriculture Technology Co., Ltd. in Beijing. To reduce the influence of structural differences between the simulation and the real robotic system on policy transfer, both policy training and transfer were performed in the end-effector action space. Specifically, end-effector pose control was adopted as a unified interface, enabling the learned policy to be effectively transferred from the simulation platform to the real robotic system. By combining simulation and real-world experiments, the effectiveness of the proposed method was further validated.

\subsection{Simulation Results and Analysis}

\begin{figure} [H]
\centering 
\includegraphics[width=0.48\textwidth]{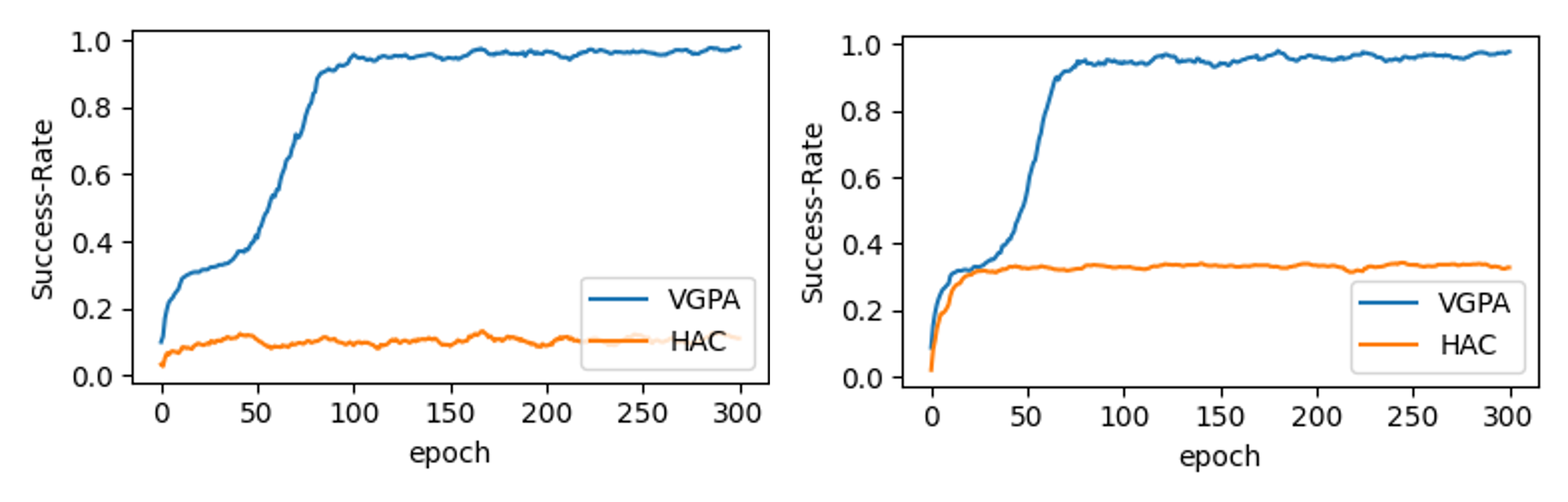} 
\caption{Success rate training curves for HAC and VGPA.} 
\label{fig:2} 
\end{figure}

1) Comparison with the HAC: We compared this approach with  Hierarchical Actor-Critic (HAC), a representative hierarchical reinforcement learning baseline. HAC was particularly suited to multi-objective tasks, as it imposed objective constraints on policies across all levels of the hierarchical structure, thereby enabling flexible sub-goal decomposition and execution.The HAC baseline was implemented under the same experimental settings described above.
As shown in Fig.~\ref{fig:2}, the proposed VGPA achieved a high-level success rate of 96.7\% and a low-level success rate of 97.54\%, demonstrating stable and efficient learning across the hierarchy. In contrast, HAC agents showed significantly lower performance, with only 7.77\% high-level success and 35.22\% low-level success. This large performance gap indicated that HAC agents were mostly limited to completing the initial obstacle separation task and struggled to accomplish the full clearance-and-grasp sequence.

\begin{figure}[H]
\centering 
\includegraphics[width=0.48\textwidth]{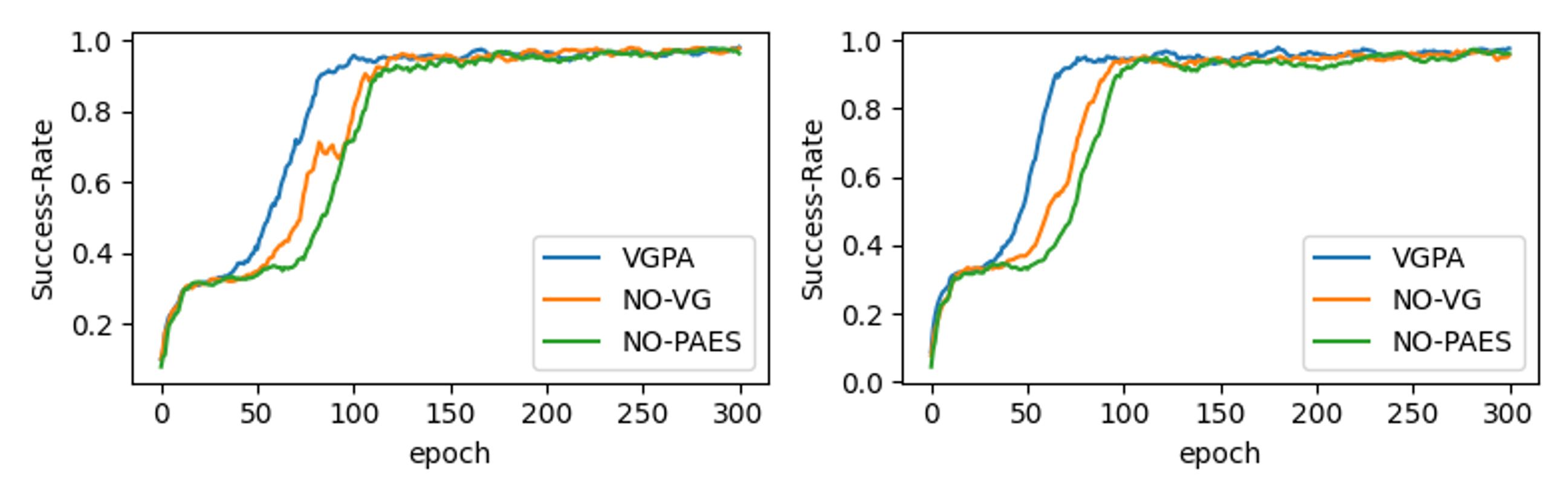} 
\caption{The ablation results for VG and PAES modules.} 
\label{fig:3} 
\end{figure}

2) Ablation studies: This section performs ablation experiments on the proposed components to validate the effectiveness of each module within the hierarchical content-driven planning VGPA framework.
The experimental results are shown in Fig.~\ref{fig:3}. Compared to the proposed VGPA method, both NO-PAES and NO-VG showed certain shortcomings. Notably, NO-VG lagged significantly behind VGPA during the early learning phase of high-level strategies, confirming that vision-guided strategies effectively assist high-level strategies in selecting the correct option, thereby substantially enhancing their learning efficiency. Concurrently, PAES mitigated instability in high-low layer training by adaptively reducing the exploration ratio and introducing an adaptive guidance mechanism, thereby accelerating the convergence of the low-level strategy. The combination of these approaches yielded a marked improvement in the overall performance of the hierarchical strategy.

\begin{figure*}[t]
\centering
\includegraphics[width=1.0\textwidth]{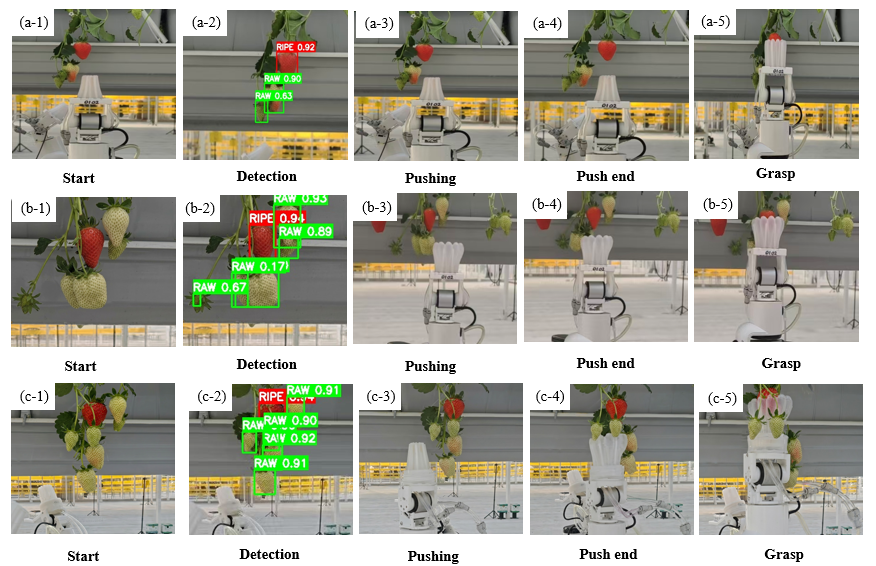}
\caption{Real-world harvesting examples under different occlusion conditions. Rows (a), (b), and (c) correspond to slight, moderate, and dense occlusion scenarios, respectively. From left to right, columns (1)--(5) illustrate the initial scene, target detection, obstacle pushing, push completion, and final grasping.}
\label{fig:real_examples}
\end{figure*}

3) Real-world Experiments: %To evaluate the effectiveness of the proposed method,  experiments were conducted on a real robotic platform at Cuihu Farm, Haidian District, Beijing, China. 
The experiments were carried out in clustered strawberry environments under three representative occlusion conditions: slight occlusion, moderate occlusion, and dense occlusion. In the slight occlusion setting, the obstacle strawberry was mainly located on one side of the target fruit, and two common cases were considered: the obstacle was positioned near the middle side of the target, or it was located in the lower-middle region beneath the target. In the moderate occlusion setting, the obstacle strawberry was mainly distributed in the lower-middle region of the target, while additional side occlusion was also present. In the dense occlusion setting, one or more obstacle strawberries were located beneath the target fruit, and lateral occlusion from both sides could also occur. For each occlusion condition, 20 different scenes were constructed, and each scene was tested in three repeated trials. Representative real-world harvesting examples under slight, moderate, and dense occlusion conditions are shown in Fig.~\ref{fig:real_examples}.

A direct picking strategy was adopted as the baseline, in which the robot directly attempted to grasp the target fruit without performing any prior obstacle-separation operation. In relatively simple scenarios, where the target fruit was sufficiently exposed and the occlusion was weak, this method could complete the harvesting task. However, under more complex occlusion conditions, it encountered significant difficulties, indicating clear limitations in handling severe occlusion and dense interference.

In contrast, the proposed obstacle-separation-based harvesting method first removed or pushed aside the fruits obstructing the target before executing the grasping action. This created a more feasible approach space for the end-effector and reduced the occlusion of the target fruit. As shown in Table~\ref{tab:realworld_comparison}, the proposed method achieved success rates of 88.3\%, 80.0\%, and 71.7\% under slight, moderate, and dense occlusion, respectively, consistently outperforming the direct picking baseline, which achieved 75.7\%, 58.3\%, and 45.0\% in the corresponding scenarios. Overall, the proposed method achieved an average success rate of 80.0\%, compared with 59.7\% for direct picking. 
The grasping time reported included the complete execution pipeline, including target recognition, motion planning, and grasp execution, and was measured from the beginning of visual recognition to the end of fruit picking. Although the average grasping time of our method (4.15~s) was longer than that of direct picking (2.93~s), the proposed strategy provided substantially higher robustness and reliability in cluttered environments. These results demonstrated that the learned policy can be effectively transferred from simulation to the real robotic system and achieves more reliable harvesting performance than direct picking in challenging clustered strawberry scenarios. 

We further compared the proposed method with a Vision-Language Model (VLM)-based method developed in our group, which used a large language model for action planning \cite{zhao5896233active}. As shown in Table~\ref{tab:realworld_comparison}, the VLM method achieved slightly higher success rates across all occlusion conditions, indicating stronger generalization. However, its execution time (7.23~s) was substantially longer than that of the proposed method (4.15~s). This suggested that, although the VLM-based method improved adaptability, the proposed RL-based method achieved a better trade-off between harvesting success and execution efficiency.

Nevertheless, several failures were still observed in the real-world experiments. First, severe occlusion affected target perception and localization. Second, variations in the size and shape of real strawberries introduced additional uncertainty during manipulation, which could also lead to harvesting failure. Third, in some complex occlusion scenarios, the current push-and-grasp strategy could swallow both the obstacle strawberry and the target strawberry at the same time during the swallowing motion, leading to harvesting failure. These observations indicate that relying solely on the push-and-grasp strategy still has certain limitations.

\begin{table}[t]
\centering
\caption{Comparison of harvesting performance under different occlusion levels}
\label{tab:realworld_comparison}
\renewcommand{\arraystretch}{1.15}
\setlength{\tabcolsep}{4pt}
\begin{tabular}{l|l|c c}
\hline
\textbf{Method} & \textbf{Scene} & \textbf{Success Rate} & \begin{tabular}[c]{@{}c@{}}\textbf{Average Grasp}\\ \textbf{Time (s)}\end{tabular} \\
\hline
\multirow{3}{*}{Direct Picking}
& Slight Occlusion   & 75.7\% & \multirow{3}{*}{2.63s} \\
& Moderate Occlusion & 58.3\% &  \\
& Dense Occlusion    & 45.0\% &  \\
\hline  
\multirow{3}{*}{Ours}
& Slight Occlusion   & 88.3\% & \multirow{3}{*}{4.15s} \\
& Moderate Occlusion & 80.0\% &  \\
& Dense Occlusion    & 71.7\% &  \\
\hline
\multirow{3}{*}{VLM}
& Slight Occlusion   & 90.0\% & \multirow{3}{*}{7.23s} \\
& Moderate Occlusion & 81.0\% &  \\
& Dense Occlusion    & 80.0\% &  \\
\hline
\end{tabular}
\end{table}

\section{CONCLUSIONS} 
This paper proposed a hierarchical reinforcement learning framework for selective strawberry harvesting in clustered environments with severe occlusion. By integrating vision-guided decision-making and adaptive exploration, the proposed method improved policy learning stability and harvesting performance. Experimental results demonstrated that the learned policy can be successfully transferred from simulation to a real robotic system and consistently outperformed the direct picking strategy under different occlusion conditions. Although the proposed simulation environment covered representative occlusion scenarios encountered in clustered strawberry harvesting, it still simplified certain characteristics of real agricultural environments, such as fruit deformation, peduncle flexibility, and complex interactions among neighboring fruits. The proposed framework provided a practical solution for selective harvesting in densely clustered strawberry environments. 

Future work will further reduce the sim-to-real gap by incorporating more realistic deformable plant models and more diverse occlusion scenarios, while improving policy transferability and refining the obstacle-separation strategy. %We also plan to introduce a dragging operation to move the target fruit to a more accessible position before grasping, thereby improving harvesting performance in complex occlusion scenarios. %In addition, comparisons with representative flat reinforcement learning algorithms will be conducted to further evaluate the effectiveness of the proposed hierarchical task decomposition.

\section{ACKNOWLEDGMENT} 
The accompanying video demonstrating both simulation and real-world experiments can be found at: \url{https://youtu.be/zHSrkjB_unE}.

% \noindent\textbf{Supplementary Material:} The accompanying video demonstrating both simulation and real-world experiments is available at \href{https://1drv.ms/v/c/e68b602748988e28/IQB3pX9798OyTLu9SzlGD2gxAZmKfrdJzuZ5mOaSnPZNrCc?e=cXCVSw}{the video link}.

% \printbibliography
\bibliographystyle{unsrt}
\bibliography{ref}

\end{document}